\documentclass[sigconf]{acmart}

\usepackage[capitalise, noabbrev]{cleveref}
\usepackage{amsmath}
\usepackage[hypcap=false]{caption}
\usepackage{subcaption}
\usepackage{multirow}
\usepackage{ctable}
\usepackage{threeparttable}

\acmSubmissionID{1042}

\begin{document}

\title{ Initial Investigation of  Kolmogorov-Arnold Networks (KANs) as Feature Extractors for IMU Based Human Activity Recognition}


\author{Mengxi Liu}
\email{mengxi.liu@dfki.de}
\affiliation{%
  \institution{German Research Center for Artificial Intelligence}
  \streetaddress{Trippstadter Str. 122}
  \city{Kaiserslautern}
  \country{Germany}
  \postcode{67663}
}

\author{Daniel Geißler}
\email{daniel.geissler@dfki.de}
\affiliation{%
  \institution{German Research Center for Artificial Intelligence}
  \streetaddress{Trippstadter Str. 122}
  \city{Kaiserslautern}
  \country{Germany}
  \postcode{67663}
}

\author{Dominique Nshimyimana}
\email{dominique.nshimyimana@dfki.de}
\affiliation{%
  \institution{German Research Center for Artificial Intelligence}
  \streetaddress{Trippstadter Str. 122}
  \city{Kaiserslautern}
  \country{Germany}
  \postcode{67663}
}

\author{Sizhen Bian}
\email{sizhen.bian@pbl.ee.ethz.de}
\affiliation{%
  \institution{ETH Zurich}
  \streetaddress{Rämistrasse 101}
  \city{Zurich}
  \country{Switzerland}
  \postcode{8092}
}

\author{Bo Zhou}
\email{bo.zhou@dfki.de}
\affiliation{%
  \institution{German Research Center for Artificial Intelligence}
  \streetaddress{Trippstadter Str. 122}
  \city{Kaiserslautern}
  \country{Germany}
  \postcode{67663}
}

\author{Paul Lukowicz}
\email{paul.lukowicz@dfki.de}
\affiliation{%
  \institution{German Research Center for Artificial Intelligence}
  \streetaddress{Trippstadter Str. 122}
  \city{Kaiserslautern}
  \country{Germany}
  \postcode{67663}
}


\begin{abstract}


In this work, we explore the use of a novel neural network architecture, the  Kolmogorov-Arnold Networks (KANs) as feature extractors for sensor-based (specifically IMU)  Human Activity Recognition (HAR). Where conventional networks perform a parameterized weighted sum of the inputs at each node and then feed the
result into a statically defined nonlinearity, KANs perform non-linear computations represented by B-SPLINES on the edges leading to each node
and then just sum up the inputs at the node. Instead of learning weights, the system learns the spline parameters. In the original work, such networks have been shown to be able to more efficiently and exactly learn sophisticated real valued functions e.g. in regression or PDE solution. We hypothesize that such an ability is also advantageous for computing low-level features for IMU-based HAR. 
To this end, we have implemented KAN as the feature extraction architecture for IMU-based human activity recognition tasks, including four architecture variations. 
We present an initial performance investigation of the KAN feature extractor on four public HAR datasets. It shows that the  KAN-based feature extractor outperforms CNN-based extractors on all datasets while being more parameter efficient.


\end{abstract}




\maketitle

\section{Introduction}

Over the last decade, sensor-based HAR has been much slower to benefit from general advances in ML techniques than areas such as computer vision and NLP. Beyond known problems such as lack of sufficient labeled training data and ambiguity of sensor signals a core issue has been the fact that many of those techniques were specifically designed to deal with the way information is encoded in images and text, which is different from the way information is encoded in sensor signals such as IMUs. In particular, Convolutional Neural Networks (CNNs) which have started the computer vision (CV) revolution and are still widely used as low-level feature extractors at the input of modern transformer systems are specifically designed around the way information is encoded in images: through a hierarchy of relative color/intensity patterns of local structures. The sensor-based HAR community has come up with the idea of organizing multimodal time series as „fake images“ to take advantage of CNN's which improved performance and became a standard technique for low-level feature extraction  \cite{ordonez2016deep} but the improvement remains much more modest than the dramatic effect CNNs had on CV. The question of what sort of network would reflect the way information is encoded in sensors, in particular motion sensors (e.g IMUs), more appropriately remains an open question.  Experience with hand-crafted, highly optimized features extractors that were state of the art before the deep learning era \cite{sargano2017comprehensive,dong2019har} suggests that (at least at a lower level) core information is often contained in time dependant, real valued functions of the inputs rather than in mere spatial patterns. 


This work investigates the use of recently proposed Kolmogorov-Arnold Networks (KAN, \cite{liu2024kan}) as feature extractors for motion sensors based HAR.  Whereas classical neural networks perform a parameterized weighted sum of the inputs at each node and then feed the result into a statically defined nonlinearity, KANs perform parameterized non-linear computations on the edges leading to each node and then just sum up the inputs at the node (see Figure \ref{fig: KAN_Arch}).  The functions that are computed on the edges are represented as B-splines with the spline parameters being the learnable parameters of the network. The core idea is that KANs are better at modeling complex real valued functions than CNNs (as they were in the original paper conceived with modeling physical systems in mind), thus better mimicking the way information is encoded in sensors, at least at the lower level.  

 From the above idea, the contribution of the paper is two folds:
 \begin{enumerate}
     \item We proposed a KAN-based low-level feature extraction architecture for IMU-based HAR, including four variation architectures.
     \item We present an initial investigation of the performance of the KAN feature extractor on four standard HAR datasets (PAMAP2, WISDM, MM-Fit, and MotionSense). It shows that the  KAN feature extractor outperforms CNN-based extractors on all datasets while being more parameter efficient.
 \end{enumerate}
Note that the purpose of this paper is not the design of the best "end-2-end" classification system but to provide an initial understanding of the potential of KANs, which are a fundamentally new type of neural networks, for sensor-based HAR, specifically low-level feature extraction from IMU signals. For this reason, we input the features produced by our extractor to a straightforward multilayer perception (MLP) with a widely used CNN-based feature extractor as a baseline rather than embed our extractor in some more advanced end-2-end architecture.   
Our core contribution is thus to provide the motivation and a stating point for follow up research to build on to improve existing and create novel end-2-end systems incorporating KAN components. 

\begin{figure*}[!t]
\footnotesize
\centering
\includegraphics[width=2.0\columnwidth]{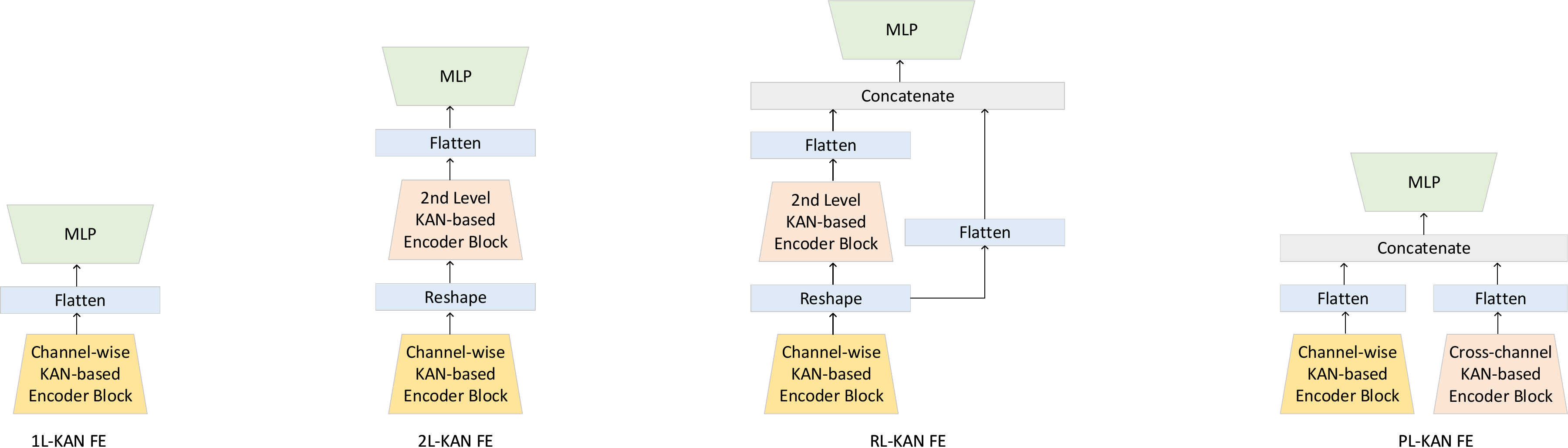}
\caption{Proposed KAN-based Feature Extractor Models (FE: Feature Extractor)}
\label{fig:KANFE_Arch_s}
\end{figure*}

\section{Related Work}
Effectively extracting low-level features from the time series is fundamental to successful sensor based HAR. 
Over the decades, feature extraction methods can be grouped into two types: handcrafted features and automatic feature extractions.
The hand-crafted methods based on human experience, which are limited in general sensor configuration, typically produce shallow features like mean, variance, and frequency, particularly useful for recognizing simple activities. 
However, they are inadequate for complex, context-aware tasks \cite{wang2019deep}, which can be better addressed by the automatic feature extraction methods based on the neural networks, like CNN \cite{taylor2010convolutional}, RNN \cite{Weerakody2021A}, and Transformer \cite{Luptáková2022Wearable, Shavit2021Boosting} based feature extractors.

Starting with the CNN architectures, such architectures are widely used in HAR due to their ability to capture local dependencies in time series data \cite{Zebin2016Human}.
Even though CNNs are traditionally utilized as a baseline comparison, recent works showcase that deep convolutional networks are still relevant and competitive \cite{Ata2022A}.
On the other hand, Recurrent Neural Networks (RNNs) and their variants, such as Long Short-Term Memory (LSTM) and Gated Recurrent Units (GRU), are particularly effective for sequential data modeling. These architectures are designed to handle temporal dependencies and can be adapted to manage irregular time series data \cite{Weerakody2021A}.
Transformer models emerged as a promising alternative next to the introduced network architectures of the HAR field.
Multiple works have proven to provide higher accuracy and better generalization across various datasets \cite{Luptáková2022Wearable, Shavit2021Boosting}.
In this work, the CNN-based feature extractor was selected as the baseline.



Recent works have utilized Kolmogorov-Arnold representations as a method for decomposing a continuous multivariate function into a hierarchical structure of functions of one variable, which can be particularly useful in machine learning applications \cite{liu2024kan}. 
The use of KANs for time series forecasting has been introduced in \cite{xu2024kolmogorov}, proposing two variants: T-KAN and MT-KAN. 
T-KAN detects concept drift and explains nonlinear relationships in time series, while MT-KAN improves predictive performance in multivariate time series by uncovering complex variable relationships. 
Experiments show that both variants outperform traditional methods in predictive accuracy and interpretability, positioning KAN as a promising tool in adaptive forecasting and predictive analytics.
KAN-based models have also been proposed in HAR tasks to achieve better results \cite{moryossef2024optimizing,liu2024ikan}, especially demonstrating promising potential in feature extraction.
The work was proposed in the Conv-KAN \cite{torch-conv-kan} to process image, which replaced the convolutional kernel with KAN consisting of a set of univariate non-linear functions to improve the feature extraction ability while keeping the same architecture as the conventional convolution.
Our work proposed the feature extractor based on multiple KANs, working as the filter blocks to extract features from time series.
In this work, we conduct an initial study of KANs as feature extractors for IMU-based Human Activity Recognition.

\section{Architecture}
\subsection{Kolmogorov-Arnold Networks }

\begin{figure}[!t]
\footnotesize
\centering
\includegraphics[width=1.0\columnwidth]{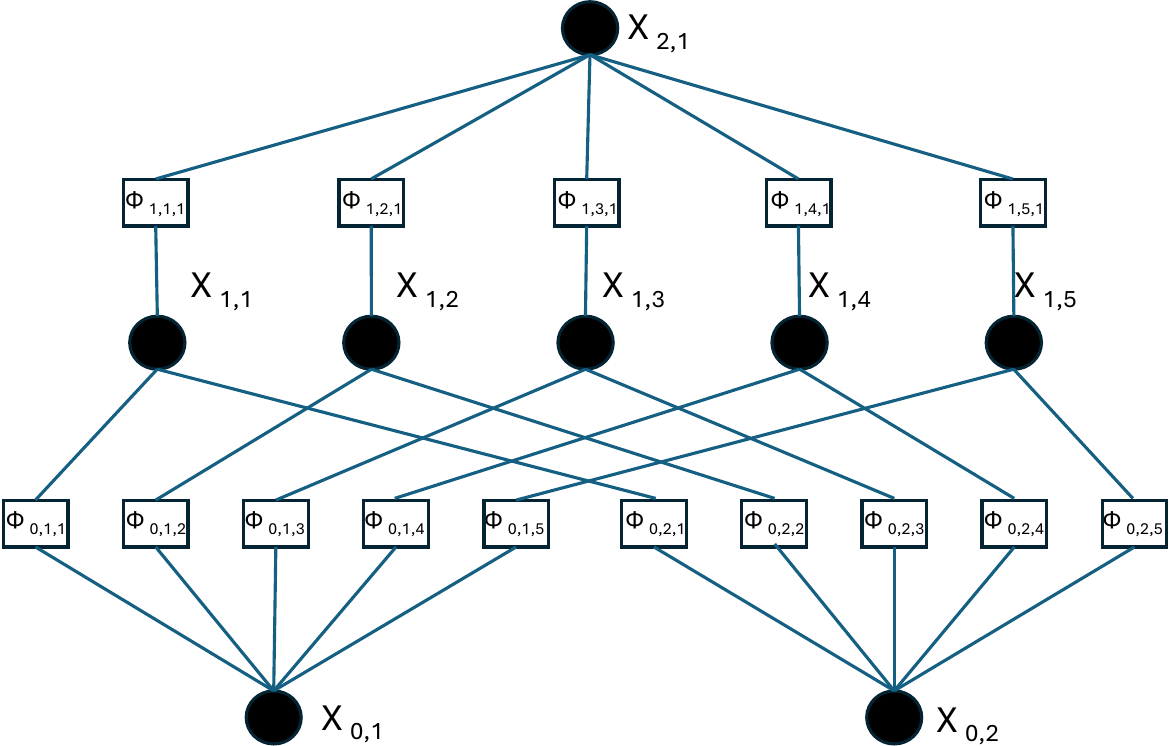}
\caption{Architecture of KAN as proposed by \cite{liu2024kan}}
\label{fig: KAN_Arch}
\end{figure}

The KAN architecture has been proposed in a very recent paper \cite{liu2024kan} as an application of the Kolmogorov-Arnold representation theorem to machine learning. The theorem states that any multivariate continuous function on a bounded domain can be represented by a finite sum of continuous functions on a single variable. 
Specifically, we have
\begin{equation}
f(x_1,x_2,....,x_n)=\sum_{i=1}^{2n+1}\Phi_i(\sum_{j=1}^{n}\phi_{i,j}(x_i))    
\end{equation}
This essentially is a two-layer neural network with the first layer having $2n+1$ neurons each connected to all inputs and the second layer having just one neuron connected to the outputs of all the $2n+1$ neurons of the first layers as shown in \cref{fig: KAN_Arch}. The big difference to a classical network is that instead of just multiplying each input on the corresponding edge with an edge-specific weight and then feeding the sum of all weighted inputs into a predefined nonlinear function at the node, we now compute an edge-specific function (which is in general nonlinear)  on each edge ($\phi_{i,j}$ in the first layer neurons and $\Phi_i$ in the second layer) and then sum the output of all edge functions in the node (with no further non linearity). Whereas in the classical network the system learns the weights, in a KAN it learns the edge functions. This is done by representing those functions as B-splines 
\begin{equation}
\phi_{i,j}=\sum_{k=k}^{degree}c_k^{i,j}B_k(x)   
\end{equation}
and the learning the spline approximation parameters $c_k^{i,j}$.  The above allows the system to learn more easily complex real valued functions as needed for example for regression or partial differential equation (PDE) solutions. Both are use cases for which original paper \cite{liu2024kan} shows KAN to significantly outperform classical architectures. The idea behind our work is based on experience that the best hand-crafted features for IMU-based HAR (which have been widely used before the deep learning systems made feature engineering obsolete) are mostly real valued functions applied to sliding windows of the signal (e.g. RMS of the individual channels, relations between values of different channels, etc). The working hypothesis (which the results presented below seem to confirm) is that a KAN can learn such features more accurately and efficiently than current feature extractors such as CNNs.

\subsection{KAN Feature Extractor Architecture}
\begin{figure}[htbp]
\footnotesize
\centering
\includegraphics[width=0.9\columnwidth]{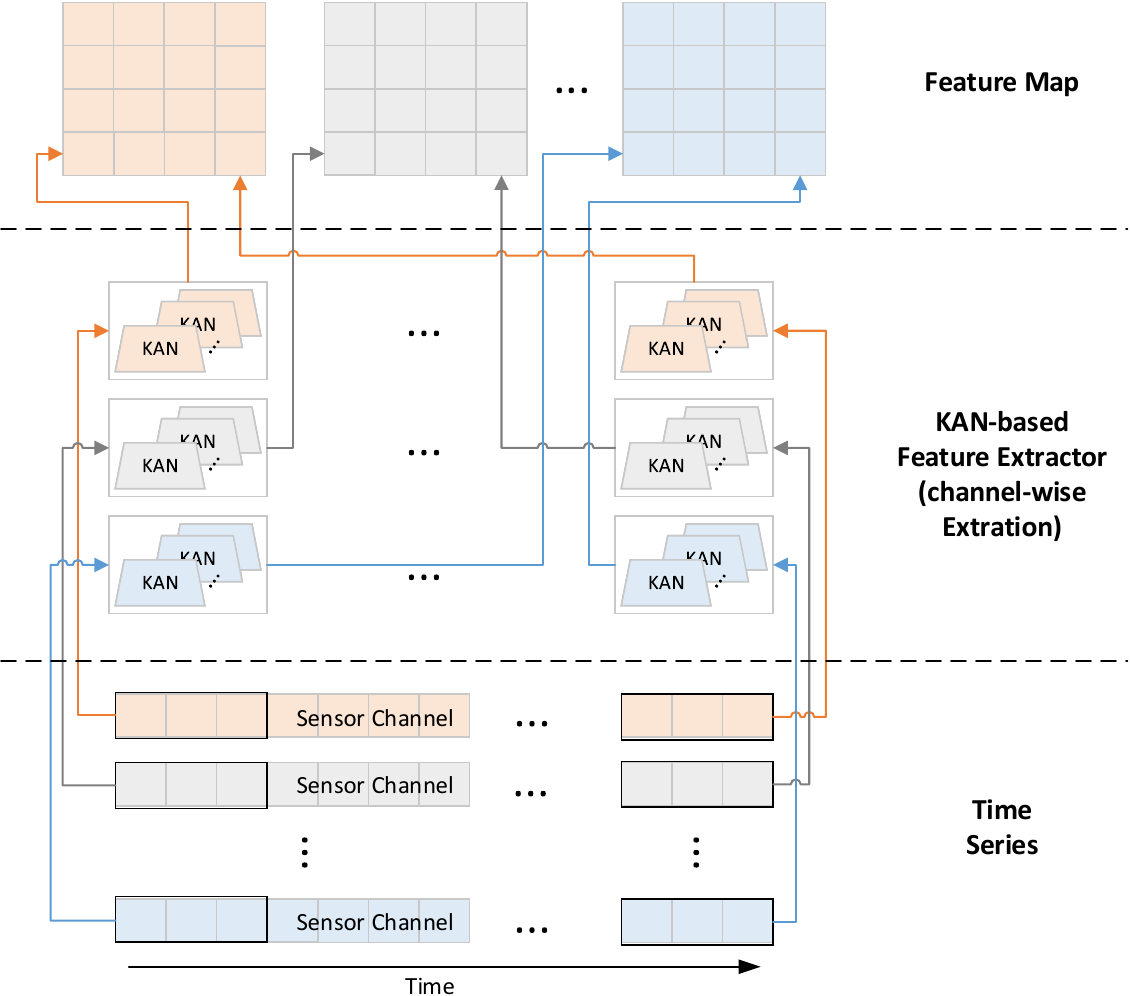}
\caption{KAN-based Feature Extractor Block (Channel-wise)}
\label{fig:KANEncoderBlock_CW}
\end{figure}

\begin{figure}[htbp]
\footnotesize
\centering
\includegraphics[width=0.9\columnwidth]{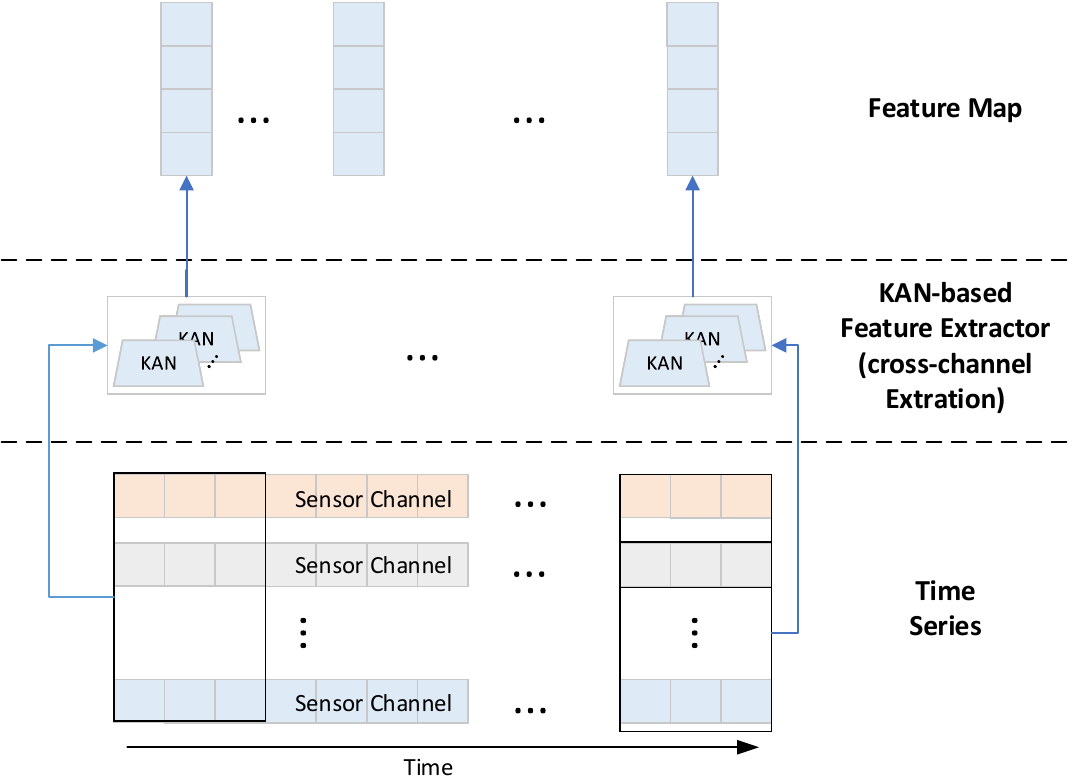}
\caption{KAN-based Feature Extractor Block (Cross-channel-wise)}
\label{fig:KANEncoderBlock_CC}
\end{figure}

The core of the proposed KAN feature extractor is shown in \cref{fig:KANEncoderBlock_CW} and \cref{fig:KANEncoderBlock_CC}. The basic building block is a KAN filter with $n_w$ inputs and one output which is identical to the basic KAN as described in the previous paragraph. Here $n_w$ is the length of a jumping window that is moved over the input frame to be classified.  We assume the frame to have $l$ windows. In analogy to the filter bank concept of CNNs we assume that for each of the windows and each of the $c$ sensor channels, there are $f$ such KAN filters, each learning a different feature on the specific channel. In its simplest version the KAN feature extractor thus outputs a $c*f$ feature matrix for each jumping window. The feature matrices are aggregated over all $l$ windows of the frame that is being classified. We refer to this as channel-wise KAN Feature Extractor Block (\cref{fig:KANEncoderBlock_CW}). In simplest version of the overall architectures the output of the block is flattened and input into a 3-layer MLP classifier, we named this architecture as \textbf{1L-KAN FE}. 
Furthermore, we consider three extensions of the basic architecture meant to explore if and how adding features computed on more than one channel in a KAN can enhance system performance (see \cref{fig:KANFE_Arch_s}).
\begin{enumerate}
    \item Addition of a second level (L2) feature extractor on top of the basic one that gets all the $c*f$ for each window as input and outputs one aggregated feature. Again we consider filter banks having $f_2$ identical KANs, each learning a different feature, we call this architecture \textbf{2L-KAN FE}. A variation of this architecture concatenates the output of L2 with the output of L1 as input to the MLP, we denote this architecture as \textbf{RL-KAN FE}. 
    \item Extension of the Feature Extraction Block through the addition of a feature extractors that get input all from all the channels in one window and output a single feature value, again using multiple $f_p$ filter banks. This we refer to as Cross Channel Feature Extraction Block (\cref{fig:KANEncoderBlock_CC}). It produces additional $l*f_p$ features which are concatenated with the input of the basic, single channel-based features before being input to the MLP. Thus the input to the MLP has  $c*f*l+l*f_p$ elements. We called this architecture \textbf{PL-KAN FE}, as shown in \cref{fig:KANFE_Arch_s}.
\end{enumerate}


\begin{figure}[!t]
\footnotesize
\centering
\includegraphics[width=1.0\linewidth]{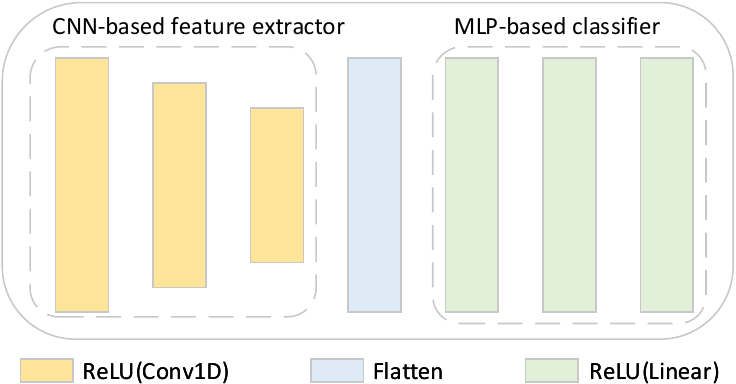}
\caption{Architecture of CNN-based Feature Extractor}
\label{fig: CNN-arch}
\end{figure}

\begin{table*}[htbp]
\renewcommand{\arraystretch}{1.0}
\footnotesize
\centering
\caption{Model Parameter (all the number is the output channel of each layer)}
\vspace{-10pt}
\label{tab:model_parameter}
\footnotesize
\begin{threeparttable}
\begin{tabular}{l c c c  c c c c c} 
\hline

\hline
Model/Layers & 1-Conv1D & 2-Conv1D & 3-Conv1D & 1-KAN & 2-KAN & 1-Linear & 2-Linear & 3-Linear\\ \hline
CNN-Enc1   & 32 & 64 & 128 & - & - & \multirow{4}{*}{256} & \multirow{4}{*}{128}& \multirow{4}{*}{class number} \\

CNN-Enc2   & 16 & 32 & 64  & - & - &  &  & \\

CNN-Enc3   & 8  & 16 & 32  & - & - &  &  &\\

CNN-Enc4   & 8  & 8  & 8   & - & - &  & &\\

\hline
1L-KAN FE    & -  & -  & -   & 5 & -  & \multirow{8}{*}{256} & \multirow{8}{*}{128}& \multirow{8}{*}{class number} \\
2L-KAN FE\_5F   & -  & -  & -   & 5 & 5  &   &  & \\
2L-KAN FE\_10F   & -  & -  & -   & 5 & 10 &  &  & \\
2L-KAN FE\_15F   & -  & -  & -   & 5 & 15 &  &  & \\
2L-KAN FE\_20F   & -  & -  & -   & 5 & 20 &  &  & \\
RL-KAN FE   & -  & -  & -   & 5 & 10 &   &  & \\
PL-KAN FE\_5F      & -  & -  & -   & 5 & 5 &   &  & \\
PL-KAN FE\_10F      & -  & -  & -   & 10 & 10 &   &  & \\

\hline
KAN-CNN-Enc0   & 5  & -  & -   & 5 & - & \multirow{3}{*}{256} & \multirow{3}{*}{128}& \multirow{3}{*}{class number} \\
KAN-CNN-Enc1   & 32 & 64 & 128   & 5 &  - &   &  & \\
KAN-CNN-Enc2   & 16  & 32  & 64   & 5 & - &   &  & \\
\hline

\hline
\end{tabular}
\end{threeparttable}
\end{table*}

\section{Experiment Configuration}

\begin{table*}[ht]
\renewcommand{\arraystretch}{1.0}
\footnotesize
\centering
\caption{Datasets for test}
\vspace{-10pt}
\label{tab:datasets}
\footnotesize
\begin{tabular}{l  l c l c c l} 
\toprule
Dataset (Year)   & Classes (Num)              & Subjects & Sensors (Channels)& Sampling Rate & Window Size & Validation Method\\ \midrule
PAMAP2 (2012) \cite{reiss2012pamap2}   & daily complex activities (13) & 8 & 3 IMUs (39)&100&200&Leave 4 subjects out\\ 
WISDM (2011)\cite{kwapisz2011wisdm} & daily locomotion activities (6)  & 36       & 1 ACC (3)& 20 &80&Leave 18 subjects out\\
MotionSense (2018)\cite{malekzadeh2018motionsense}& daily locomotion (6) & 24 & 1 smartphone in the pocket (12)&50&200&Leave 12 subjects out\\
MM-Fit (2020)\cite{stromback2020mmfit}& sports activities(11) & 10 & 5 IMUs (24)&50&200&Leave 5 subjects out\\
\bottomrule
\end{tabular}
\end{table*}

\begin{figure*}[ht]
\footnotesize
\centering
\includegraphics[width=1.0\linewidth]{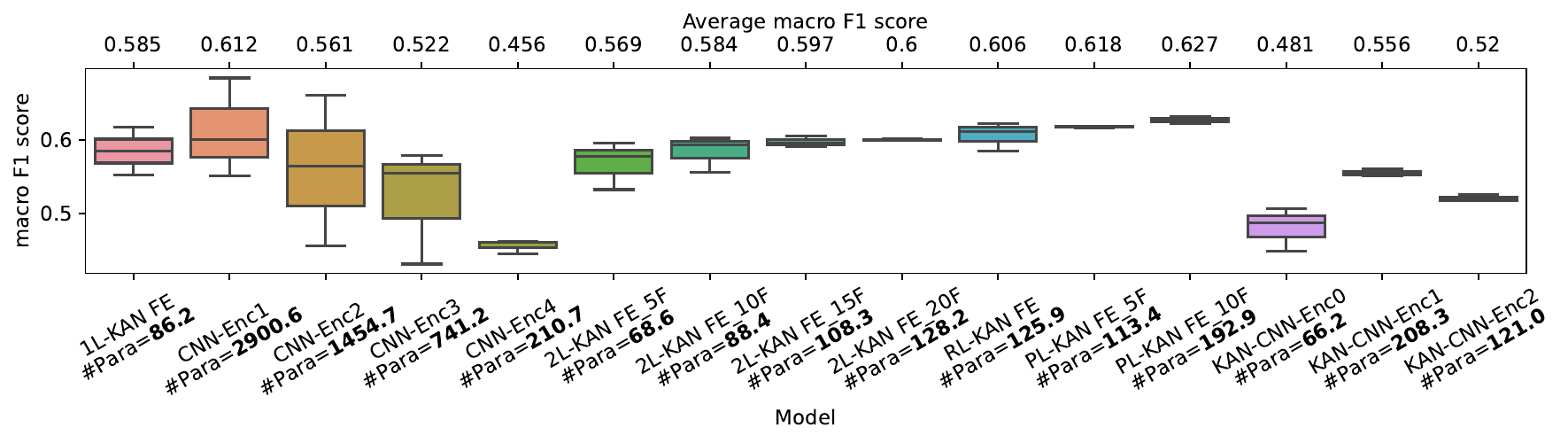}
\caption{Test result on WISDM dataset (sensor position: Pocket)}
\label{fig: WISDM_body}
\end{figure*}
\begin{figure*}[ht]
\footnotesize
\centering
\includegraphics[width=1.0\linewidth]{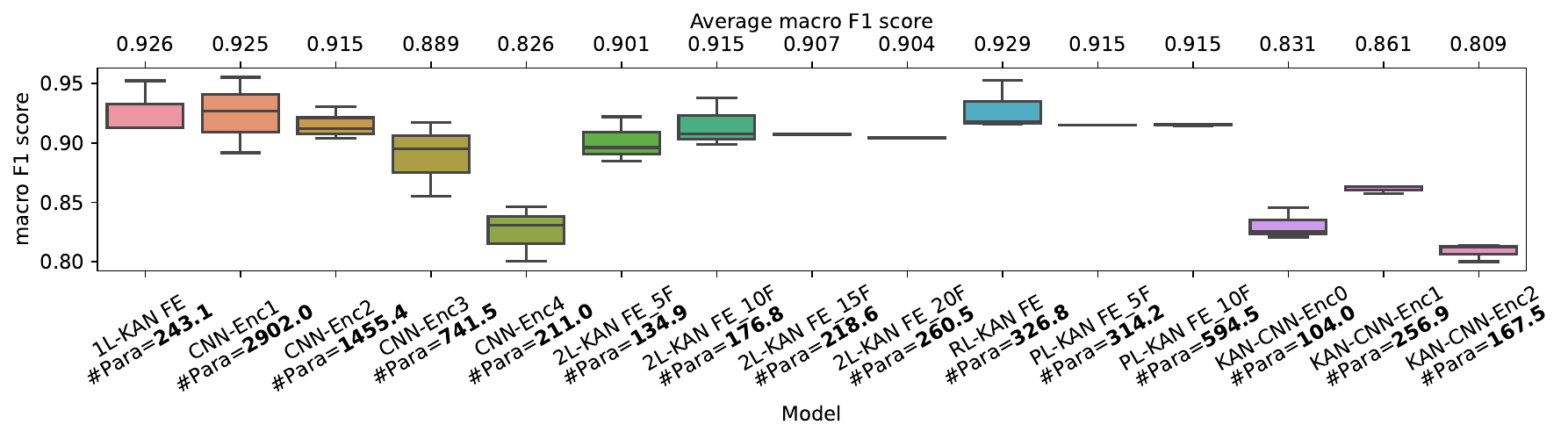}
\caption{Test result on MotionSense dataset (sensor position: Pocket)}
\label{fig: MOTIONSENSE_body}
\end{figure*}
\begin{figure*}[ht]
\footnotesize
\centering
\includegraphics[width=1.0\linewidth]{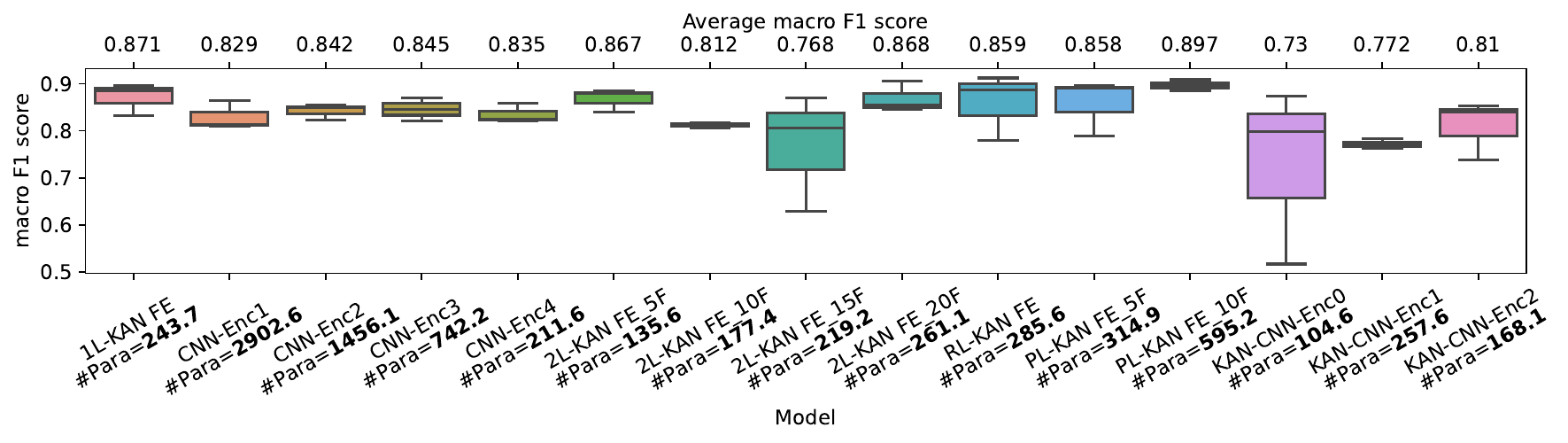}
\caption{Test result on MM-Fit dataset (sensor position: Wrists)}
\label{fig: MMFIT_wrist}
\end{figure*}
\begin{figure*}[ht]
\footnotesize
\centering
\includegraphics[width=1.0\linewidth]{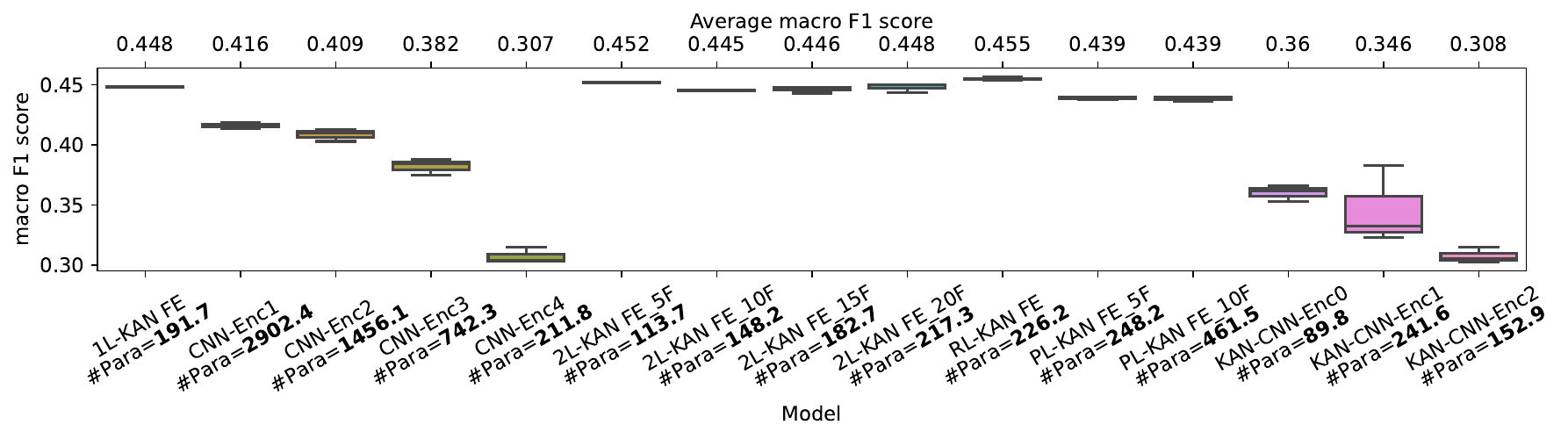}
\caption{Test result on PAMAP2 dataset (sensor position: wrists)}
\label{fig: PAMAP2_wrist}
\end{figure*}
\begin{figure*}[ht]
\footnotesize
\centering
\includegraphics[width=1.0\linewidth]{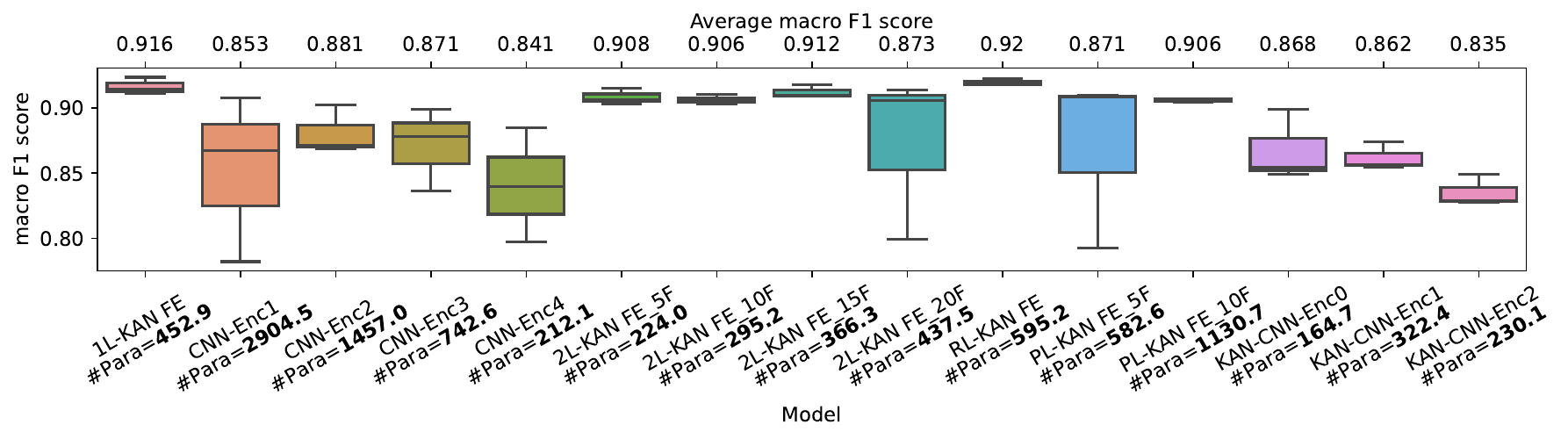}
\caption{Test result on MM-Fit dataset (sensor position: around body)}
\label{fig: MMFIT_body}
\end{figure*}
\begin{figure*}[ht]
\footnotesize
\centering
\includegraphics[width=1.0\linewidth]{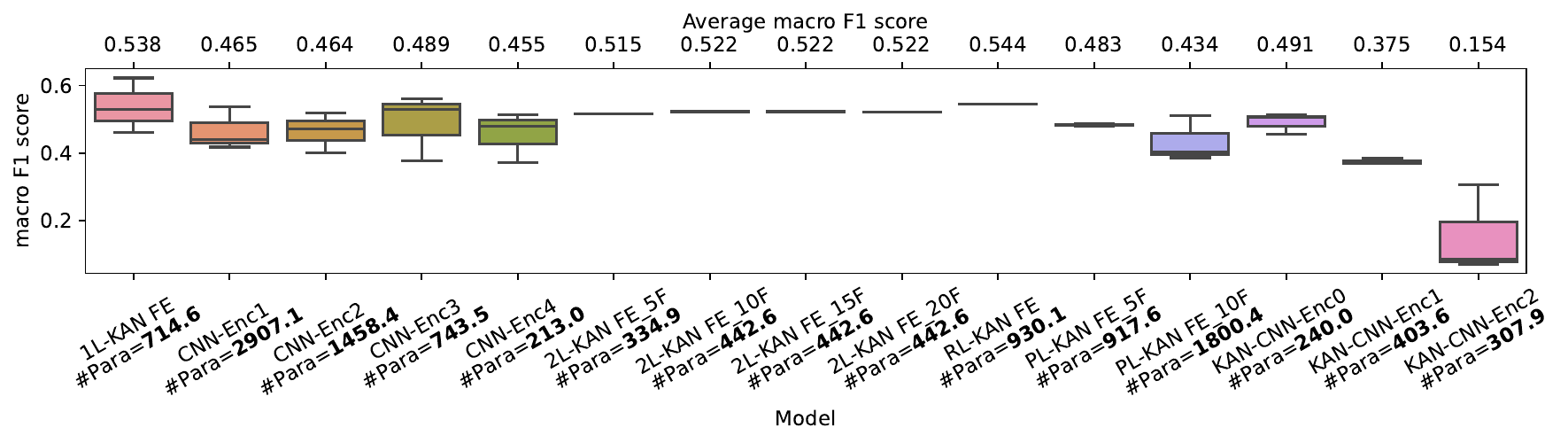}
\caption{Test result on PAMAP2 dataset (sensor position: around body)}
\label{fig: PAMAP2_body}
\end{figure*}

For the performance evaluation, we rely on four public sensor-based HAR datasets that are widely used in the community:  WISDM \cite{kwapisz2011wisdm}, MotionSense \cite{malekzadeh2018motionsense}, MM-Fit \cite{stromback2020mmfit}, and PAMAP2 \cite{reiss2012pamap2}. the datasets vary in the number of sensor channels (from WISDOM having just 3 acceleration channels to MM-Fit having 5 IMUS) and the complexity of the recognition tasks (with WISDOM and Motionsense dealing with simpler modes of locomotion problem and MM-fit and PAMAP2 looking at more complex activities (sports exercises, broader everyday life activities).  
Summary of the information is listed in \cref{tab:datasets}. 
We perform a leave-x-subject-out cross-validation, half of the subjects were selected as the test dataset, and the rest of the subjects were used as the training dataset.
The instances with a window size of 100 were generated by a sliding window approach with a slide step of 50, and a normalization was applied to each instance.
In addition, we test another two datasets derived from the PAMAP2 and MM-Fit, which only include the data from the IMU sensors on the wrists.

As a baseline a  CNN-based feature extraction model commonly used in sensor based HAR for the low level feature processing was selected as shown in  \cref{fig: CNN-arch}. It consists of three one-dimensional convolutional layers and an MLPs with one hidden layer as the classifier.
The kernel size of all convolutional layers was configured as five and the stride step is one.
The number of output channels of each convolutional layer is shown in \cref{tab:model_parameter}, we tested the CNN-based feature extraction model with four different configurations in different output channels. 
The features extracted from CNN layers were flattened and then input to the classifiers.
The output channel number of all MLPs remains the same in all experiments except the last layer, which is the same as the number of classes, the input channel of the first linear layer was varied to adapt to the different number of input features.

Based on the proposed channel-wise KAN-based feature extractor block shown in \cref{fig:KANEncoderBlock_CW} and the cross-channel-wise KAN-based feature extractor block shown in \cref{fig:KANEncoderBlock_CC}, four proposed KAN-based models shown in \cref{fig:KANFE_Arch_s} were tested in this work, such as \textbf{1L-KAN FE}, \textbf{2L-KAN FE}, \textbf{RL-KAN FE} and \textbf{PL-KAN FE}.
To investigate the effect of the number of filters of the KAN-based feature extractor, the feature numbers, including 5,10,15,20, in the channel-cross-wise KAN-based feature extractor block from the model \textbf{2L-KAN FE} were selected to test. The parameters of these models are listed in \cref{tab:model_parameter}.

In addition, we also tested additional models,\textbf{KAN-CNN-Enc}, to investigate whether the combination of KAN-based feature extractor and CNN-based feature extractor can improve the classification performance, where we used the channel-wise KAN-based feature extractor to extract the features from raw time series and then input the extracted features to convolutional layers to extract deeper features, followed by classifiers.

All the models were trained by 100 epochs with an early stop configuration, whose patience was set as 20, the sum of Accuracy and macro F1 score were used as the stop metric. 
The cross-entropy loss function and SGD (Stochastic gradient descent) optimizer with a learning rate of 0.01 and a momentum of 0.9, and 1024 batch size were used during the training process.
In the KAN-based feature extractor block, we set the grid point as 5 and the grid range from 0 to 1, the piecewise polynomial order of splines was set as 3. The other hyperparameters in KANs were kept as default. 

The macro F1 score and the model size are selected as the metrics to evaluate the performance of the proposed KAN-based feature extractor.
All the experiments were repeated three times.
The average macro F1 score was reported in this work.

\section{Results}
The results are shown in Figures \ref{fig: WISDM_body} to \ref{fig: PAMAP2_body}. For the two datasets with simpler modes of locomotion-related activities and a single sensor (WISDOM, Motionsense) the best KAN encoder performs slightly better (for Motionsense within noise margin 0.004=0.4\% better and for WISDM  0.014=1.4\%).  In both cases the number of parameters of the KAN-based system is an order of magnitude (around 10 times for Motionsense and around 20 times for WISDM) smaller. For PAMAP2 and MMFit, which contain more complex activities the best KAN encoder clearly outperforms all the CNN encoders both for the complete sensor setup and the experiment with wrist-only sensor (by between 3 and 5\%) while still retaining the smaller model size advantage (although by a smaller factor of around 5).  Within the KAN architectures, the simplest channel-wise 1L KAN encoder,  the RL-KAN-FE which runs the channel-wise and cross-channel extraction in parallel to the Level 2 information and the PL-KAN that has no L2 encoding and uses cross-channel encoder output concatenated with the individual channel encoders perform best (with the RL-KAN being best overall). This confirms the intuition that KANs are better at estimating real valued function focused features at lower levels than on more combinatorial higher-level analysis which is better left to the MLP. 
In most cases, the addition of a CNN extractor on top of the KAN performs worse the either the corresponding KAN or the corresponding CNN. We hypothesize that this is due to the complexity of training a mixed KNN/CNN network. KANs are generally slower to train than conventional networks and use a modified optimization strategy as described in the original paper (\cite{liu2024kan}).

\section{Discussion and Conclusion}
The results presented above clearly show that KAN extracted low-level features containing more relevant information about motion-related activities than CNN encoders that are widely used as the initial stage of feature extraction in most current systems. 
It also shows that KAN-based feature extractors have a better trainable parameter efficiency than CNN encoders when they obtain similar performance.
The main disadvantage of KANs is the slower training speed. In addition, training more complex mixed (KAN/traditional) networks poses a challenge as also illustrated by the poor performance of mixed KAN/CNN encoders. 

The main limitation of the study is the fact that we focused on the information extracted by the KAN/CNN encoders without investigating their embedding in state-of-the-art recognition systems (e.g. transformers). We believe that for an initial study it is the right approach as it more directly exposes the performance of the KAN encoders as such. However, to fully establish the value of t KANs for HAR complete end-2-end architectures need to be developed and evaluated as a net step. 
In the future, we also aim to investigate using approaches such as contrastive learning to pre-train KAN encoders and then freeze their weights when training the higher-level network on top. The investigation of both more complex KAN architectures such as deeper individual KAN encoders will be another topic.





\bibliographystyle{ACM-Reference-Format}
\bibliography{sample-base}
\end{document}